\documentclass[graybox]{svmult}

\usepackage{type1cm}        % activate if the above 3 fonts are
\usepackage{makeidx}         % allows index generation
\usepackage{graphicx}        % standard LaTeX graphics tool
\usepackage{multicol}        % used for the two-column index
\usepackage[bottom]{footmisc}% places footnotes at page bottom

\usepackage{newtxtext}       % 
\usepackage[varvw]{newtxmath}       % selects Times Roman as basic font

\usepackage{tikz}
\usetikzlibrary{arrows.meta,positioning,fit,calc,shapes.misc,backgrounds}

\makeindex             % used for the subject index
\begin{document}

\title*{Chapter 3: Architectures for Building Agentic AI}
\author{Slawomir Nowaczyk\orcidID{0000-0002-7796-5201}}
\institute{Slawomir Nowaczyk \at Center for Applied Intelligent Systems Research, Halmstad University, Sweden \email{slawomir.nowaczyk@hh.se}}
\maketitle

\abstract{This chapter argues that the reliability of agentic and generative AI is chiefly an architectural property. We define agentic systems as goal-directed, tool-using decision makers operating in closed loops, and show how reliability emerges from principled componentisation (goal manager, planner, tool-router, executor, memory, verifiers, safety monitor, telemetry), disciplined interfaces (schema-constrained, validated, least-privilege tool calls), and explicit control and assurance loops. Building on classical foundations, we propose a practical taxonomy—tool-using agents, memory-augmented agents, planning and self-improvement agents, multi-agent systems, and embodied or web agents—and analyse how each pattern reshapes the reliability envelope and failure modes. We distil design guidance on typed schemas, idempotency, permissioning, transactional semantics, memory provenance and hygiene, runtime governance (budgets, termination conditions), and simulate-before-actuate safeguards.}

\section{Introduction: purpose, scope, and architecture reliability}
This chapter surveys architectural choices for building agentic AI systems and analyses how those choices shape reliability. Our central claim is straightforward: \textit{reliability is, first and foremost, an architectural property}. It emerges from how we decompose a system into components, how we specify and enforce interfaces between them, and how we embed control and assurance loops around the parts that reason, remember, and act. Individual models matter, but without the right architectural scaffolding, even state-of-the-art models will behave inconsistently, be impossible to audit, and prove fragile in the face of novelty.
\begin{tikzpicture}[remember picture,overlay]
\node[fill=white, text width=12cm] (b) at (4,18){\large\color{blue}This is a preprint of a chapter accepted for publication in \textit{Generative and Agentic AI Reliability: Architectures, Challenges, and Trust for Autonomous Systems}, published by Springer Nature.};
\end{tikzpicture}

\textbf{Agentic AI} in this book denotes systems that pursue goals over time by \textit{deciding} what to do next, \textit{selecting and using tools, consulting and updating memory}, and \textit{interacting with their environment} under constraints. An agent is not merely a predictor; it is a decision-maker in a closed loop. It observes, plans (or at least chooses), acts, and learns, typically under uncertainty and partial observability. \textbf{Generative AI} refers to models that synthesise content—text, code, images, plans, or intermediate representations—often serving as the reasoning substrate inside the agent, or providing artefacts (queries, programs, simulations, explanations) that other components execute or verify. In modern systems, \textit{generative models} supply the \textit{policy} (how to reason and propose actions), while the \textit{agentic architecture} supplies the \textit{machinery} (how proposals are validated, enacted, bounded, and recorded).

Understanding the relation of \textbf{Agentic GenAI} with \textbf{classic autonomous agents} is crucially important to avoid reinventing the wheel: many key concepts have been studied for a long time and are relatively well-understood today; however, the nature of GenAI also brings up challenges that are completely novel and require rethinking of what was believed to be known. Traditional reactive, deliberative, or BDI (belief-desire-intention) architectures offer theoretically-founded and crisp notions of concepts such as beliefs, goals, plans, and intentions, with clear control loops and explicit world models. Modern agentic systems often replace hand-engineered reasoning with neural-network-based foundation models. These models, trained on huge amounts of diverse data, vastly increase the flexibility and breadth of competence, but also introduce uncertainty in reasoning steps and tool usage. In this chapter, we retain the useful discipline of the classic view—explicit state, goals, plans, commitment strategies, and monitoring—while acknowledging that parts of the pipeline (e.g., plan generation or hypothesis formation) may be implemented by generative models. That reconciliation is precisely where architecture earns its keep.

This book is not intended as yet another broad introduction to Agentic GenAI; instead, we put these recent developments in the specific context of \textbf{reliability}. By reliability, we mean \textit{the consistent achievement of intended outcomes under stated conditions, within acceptable bounds of safety, security, data protection, and resource usage, and with evidence that failure modes are known, contained, and recoverable}. For agentic AI, this encompasses much more than just model accuracy. It includes correct tool invocation, bounded action sequences, resistance to manipulation, predictable latency and cost, graceful degradation, auditability, and human-override paths. Architectures make these properties tractable by: (i) isolating \textit{responsibilities} in modules whose contracts we can reason about; (ii) interposing \textit{validators} and \textit{verifiers} between reasoning and action; (iii) \textit{constraining} authority and side-effects through permissioned tool interfaces; and (iv) \textit{instrumenting} the system so that internal state, decisions, and outcomes are observable and replayable.

Rather than hinging on a single mechanism, system-level reliability is shaped by the interaction of a few foundational architectural choices. In practice, three mutually reinforcing design choices determine how agentic systems behave under stress: how we decompose functionality, how the parts communicate and are constrained, and how their behaviour is supervised at run-time.

\textbf{Componentisation}. Separating the functionality, such as perception, memory, planning, tool routing, execution, verification, and oversight, confines faults to well-defined boundaries and limits their blast radius. Clear responsibilities make defects diagnosable and upgrades safe: a verifier can be strengthened without disturbing the planner; an execution gateway can be hardened without touching memory logic. Componentisation also enables staged deployment (mock tools, simulators, or read-only modes first) and offers natural choke points for safety checks.

\textbf{Interfaces and contracts}. Primary means to tame open-ended model behaviour are typed and schema-validated messages; explicit capability scopes for tools; idempotent and (where feasible) transactional semantics; rate/authority limits. All of these convert free-form model outputs into predictable, auditable actions. Interfaces extend to memory: retrieval must carry provenance and freshness guarantees; long-term stores need retention, compaction, and contamination controls. Good contracts enable the system to act deterministically when safe and refuse when not, transforming ambiguous proposals into either valid commands or actionable error reports.

\textbf{Control and assurance loops}. Monitors compare planned with observed behaviour; critics and verifiers check factuality, policy compliance, and safety invariants; supervisors enforce budgets, escalation rules, and termination criteria; fallbacks provide safe modes of operation when assumptions fail. These loops supply the governing feedback around generative components, preventing small reasoning slips from cascading into hazardous sequences and ensuring graceful degradation under uncertainty.

Taken together, these choices turn a powerful but free-form reasoning engine into a bounded, observable, and governable system. In the remainder of this section, to make these ideas concrete, we illustrate how they play out in the \textit{running example} of a tool-using diagnosis agent operating in a safety-critical environment. Imagine a fleet operator responsible for electric power systems in autonomous service vehicles. The agent’s mission is to triage anomalies, recommend mitigations, and, within a narrow envelope, execute pre-approved actions that reduce risk and downtime.

The agent comprises: a \textbf{Goal Manager} (ingesting alerts and operator intents), a \textbf{Perception and Retrieval} layer (querying telemetry stores and maintenance logs), a \textbf{Planner} (often a generative model producing hypotheses, tests, and action candidates), a \textbf{Tool Router} (mapping abstract actions to concrete, permissioned tools: telemetry queries, digital-twin simulation, firmware status, dispatch scheduling), an \textbf{Execution Gateway} (schema validation, pre-condition checks, simulators-before-actuators, idempotency tokens), a \textbf{Verifier/Critic} (analysing proposed explanations and commands against rules, limits, and known hazards), a \textbf{Memory subsystem} (short-term scratchpads for the current case, long-term episodic/semantic stores with provenance), and a \textbf{Safety Supervisor} (budgets, escalation, and safe-halt rules). All interactions generate structured logs that are stored in an immutable audit trail.

A typical episode unfolds as follows. An over-temperature alert arrives from Vehicle V. The Goal Manager formulates a diagnosis task. The Planner drafts a hypothesis: recent fast-charge sessions combined with ambient heat may have accelerated cell imbalance. It proposes a sequence: retrieve finer-grained thermal maps; run a digital-twin simulation under current boundary conditions; if the risk exceeds the threshold, schedule a derated operating mode and prompt the operator to plan a service stop. The Verifier checks that the hypothesis is consistent with known failure modes and that the proposed tool calls are within policy. The Tool Router prepares calls with fully specified schemas; the Execution Gateway validates parameters and runs the simulation in a sandbox. If the predicted risk exceeds the configured limit, the Supervisor authorises a reversible derating command and raises a priority ticket. If any check fails—due to a schema mismatch, missing data, or conflicting evidence—the Supervisor triggers a safe-halt path: no actuation, a concise explanation, and immediate escalation to a human. Every step is logged, and the episode is later replayable for audit and improvement.

\begin{figure}[t]
    \centering
\begin{tikzpicture}[
  font=\small, >=Latex,
  node distance=12mm and 8mm,
  block/.style={draw, rounded corners, minimum height=8mm, align=center, text width=28mm},
  smallblk/.style={draw, rounded corners, minimum height=7mm, align=center, text width=24mm},
  tag/.style={draw, rounded corners, minimum height=7mm, align=center, text width=24mm},
  band/.style={draw, dashed, rounded corners, inner sep=5pt},
  note/.style={align=left, inner sep=1pt}
]

% === Vertical spine (narrow) ===
\node[block] (goal)    {Goal manager:\\\footnotesize intent, constraints, normalisation};
\node[block, below=of goal] (plan)   {Planner:\\\footnotesize reasoning,\\ decomposition};
\node[block, below=of plan] (router) {Tool router:\\\footnotesize schema-based I/O, least privilege};
\node[block, below=of router] (gate) {Execution gateway:\\\footnotesize rate limits, retries,\\ idempotency};
\node[block, below=of gate, text width=32mm, minimum height=10mm] (surface) {\textbf{Action surface:}\\ \footnotesize APIs/tools};
% : Search\,|\,Code\,|\,DB\,|\,HTTP\,|\,FS\,|\,Chat

\draw[->] (goal) -- node[left]{task spec} (plan);
\draw[->] (plan) -- node[left, xshift=1mm]{plan / actions} (router);
\draw[->] (router) -- node[left]{actuate} (gate);
\draw[->] (gate) -- (surface);

% Observations / feedback (keep tight at top-left)
\node[smallblk, left=12mm of goal, yshift=+12mm] (obs) {Observations and feedback};
\draw[->] (obs) -- (goal);

% === Memory: stacked at left to keep width compact ===
\node[smallblk, left=12mm of plan, yshift=+12mm] (wm)   {Working memory};
\node[smallblk, below=17mm of wm] (epis) {Episodic memory};
\node[smallblk, below=15mm of epis]  (sem)   {Semantic / vector store};

\node[band, fit={(wm) (epis) (sem)}] (membox) {};
\node[anchor=south west, note] at ($(membox.north west)+(1pt,2pt)$) {\footnotesize \textbf{Memory}};

% Memory links (short, to limit width)
\draw[->] (goal)     -| (wm);
\draw[->] (wm.south) |- ($(plan.west)+(0pt,5pt)$);
\draw[->] ($(plan.west)-(0pt,5pt)$)   -| (epis);
\draw[->] ($(router.west)-(0pt,5pt)$) -| (sem);
\draw[->] (epis)   |- ($(router.west)+(0pt,5pt)$);

% === Assurance & control band (wrap only the spine) ===
\node[band, fit={(goal) (plan) (router) (gate)}] (assure) {};
\node[anchor=south west, note] at ($(assure.north west)+(1pt,2pt)$) {\footnotesize \textbf{Assurance \& control}};

% Assurance tags (kept tight at right side)
\node[tag, right=12mm of gate] (bud)  {Budgets \& termination};
\node[tag, right=12mm of router]   (sim)  {Sandbox \& simulator};
\node[tag, right=12mm of plan, yshift=-3mm]   (ver)  {Verifiers \&\\ critics};
\node[tag, right=12mm of goal, yshift=12mm] (safety) {Safety supervisor};

% Short hook arrows into the spine
\draw[->] (ver) -- node[above, xshift=20mm, yshift=-2mm]{\footnotesize validate-before-actuate} ($(plan.south)!0.75!(router.north)$);
\draw[->] (sim) -- node[above, xshift=20mm, yshift=-2mm]{\footnotesize simulate-before-actuate} ($(router.south)!0.75!(gate.north)$);
\draw[->] (bud) -- node[above, xshift=20mm, yshift=-2mm]{\footnotesize step/latency limits} ($(gate.south)!0.75!(surface.north)$);
\draw[->] (surface.east) .. controls +(45mm,0mm) .. (safety.south east);
\draw[->] (safety) -- (goal);

% Observability / audit (single sink at bottom to avoid lateral spread)
\node[smallblk, below=20mm of sem, xshift=-2mm] (audit) {Observability\\ and audit log};
%%\foreach \n in {goal,plan,router,gate,wm,epis,sem,surface}
%%  \draw[-{Latex[length=2mm]},dashed] (\n.south east) .. controls +(5mm,-3mm) and +(-5mm,3mm) .. (audit.north);
\draw[->,densely dotted] (goal.south west) .. controls +(-6mm,0mm) .. (audit.north east);
\draw[->,densely dotted] (plan.south west) .. controls +(-6mm,0mm) .. (audit.east);
\draw[->,densely dotted] (router.south west) .. controls +(-6mm,0mm) .. (audit.east);
\draw[->,densely dotted] (gate.south west) .. controls +(-6mm,0mm) .. (audit.south east);
\draw[->,densely dotted] (wm.south west) .. controls +(-6mm,0mm) .. (audit.north west);
\draw[->,densely dotted] (epis.south) .. controls +(-6mm,0mm) .. (audit.north);
\draw[->,densely dotted] (sem.south) .. controls +(6mm,0mm) .. (audit.north);
\draw[->,densely dotted] (surface.south west) .. controls +(-6mm,0mm) .. (audit.south);

% Optional compact caption (remove if embedding in a figure environment)
%\node[align=center, below=5mm of audit] (cap) {\footnotesize Reliability is earned architecturally via typed interfaces, least privilege,\\
%assurance hooks (validators, simulation), budgets/termination, safety, and observability.};

\end{tikzpicture}
    \caption{Reliability is earned architecturally via typed interfaces, least privilege, assurance hooks (validators, simulation), budgets/termination, safety, and observability.}
    \label{fig1}
\end{figure}
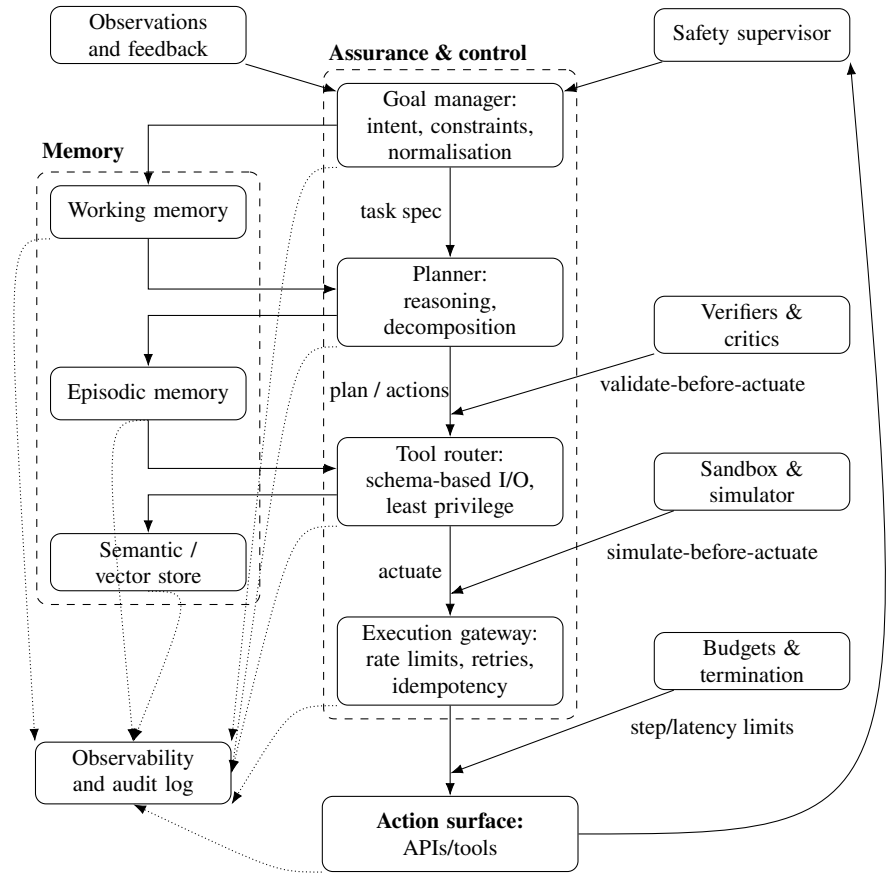

Let us examine precisely where the architecture delivers reliability in this scenario. \textbf{Containment and least authority}: the agent cannot call arbitrary tooling. Each tool is permissioned with narrow scopes (read-only telemetry; simulation only; actuation only with reversible, capped parameters). This prevents a reasoning slip from becoming a safety incident. \textbf{Validators before actuators}: plans and tool arguments are validated against schemas, pre-conditions, invariants, and policy. Invalid or unsafe proposals fail fast, preserving system integrity and clarifying where the defect lies (planner vs environment vs interface). \textbf{Assured fallbacks and graceful degradation}: when uncertainty is high or checks fail, the Supervisor enforces a conservative default (monitor-only, derate-only, or escalate). Reliability is thus not the absence of failure, but the boundedness of failure modes. \textbf{Observability and auditability}: structured traces of observations, thoughts, plans, validations, and actions enable root-cause analysis, targeted retraining, and regulatory evidence. Without this, improvement is guesswork, and accountability is weak. \textbf{Memory hygiene}: the memory subsystem separates transient scratchpads from durable knowledge; it records provenance, enforces retention policies, and mitigates contamination. This preserves reasoning quality over long horizons. \textbf{Cost and latency governance}: budgets and termination criteria bound resource use and prevent unending tool chains or self-critique loops, a practical dimension of reliability often overlooked in purely model-centric discussions.

This running example is intentionally generic. It applies, with minor changes, to data-centre operations, industrial robotics, or clinical decision support: swap the tools, adjust the policies, retain the architectural guarantees. Throughout the chapter, we will return to this agent to illustrate how specific design patterns—tool use, memory augmentation, multi-agent protocols, verification layers—alter the reliability envelope.

Our scope for the remainder of this chapter, then, is the \textbf{taxonomy of agentic architectures} (from BDI-style hybrids to tool-using, memory-augmented, and multi-agent systems), the \textbf{building blocks and interfaces} that make them dependable, and the \textbf{design-time and run-time controls} that govern behaviour. Out of scope are detailed algorithms for every model class; we discuss models only insofar as they affect architectural choices and assurance. The aim is to equip the reader with principled templates and checklists, ensuring reliability is designed in from the first diagram, rather than bolted on after the first incident.

\section{Classical foundations}

Classical agent architectures provide the control abstractions that modern agentic systems still rely on, even when parts of the stack are implemented with generative models. They clarify where decisions happen, what information is exchanged, and how behaviour is governed over time—precisely the aspects that determine reliability.

\subsection{Reactive, deliberative, and hybrid architectures}
\textbf{Reactive architectures} map perceptions directly to actions through hand-coded condition–action rules or learned policies. Their strength is low latency and robustness to minor distributional shifts; their weakness is brittleness when tasks require look-ahead, long-horizon credit assignment, or reasoning about hidden state. Failure modes typically arise from: \textit{myopic responses}, i.e., lack of explicit planning, leading to oscillations or unsafe reflexes; \textit{partial observability}, where the same observation demands different actions depending on unobserved context; \textit{unmodelled delays}, where actuation lags make “instant” responses inappropriate.

\textbf{Deliberative architectures} maintain explicit models of the world and use search or planning to choose actions (e.g., goal decomposition, symbolic planners). They excel at explainability and goal consistency but can suffer from \textit{latency} and \textit{model mismatch}. Typical failures include: \textit{stale plans} turning unsuitable under rapidly changing environments; \textit{model incompleteness}, where the real-world violates planner assumptions; \textit{computational overrun}, where planning cannot complete within operational deadlines.

\textbf{Hybrid architectures} combine the two approaches: reactive layers handle tight control loops, while deliberative layers provide goals, constraints, and re-planning. Hybrids remain the most practical template for agentic AI because they separate concerns: fast safety where the action is; slow reasoning in supervisory layers. Failures usually stem from coordination faults—e.g., unclear authority between layers, or missing hand-off criteria—rather than from either layer alone. Accordingly, reliability depends on: \textit{well-defined interfaces}, what context the reactive layer requires and what guarantees it gives back; \textit{escalation rules}, when to invoke deliberation and when to fall back to safe defaults; \textit{time budgets and termination criteria} for planning and re-planning.

\subsection{Belief–Desire–Intention to structure behaviour}

The BDI model~\cite{BDI} frames agency as the management of Beliefs (informational state), Desires (admissible goals), and Intentions (adopted plans/commitments). Two ideas in particular make BDI enduring. The first is \textbf{commitment strategies}, as agents do not re-plan on every perturbation; they stick to intentions while conditions hold, because commitment creates predictability and reduces thrashing. The second is \textbf{intention revision}, namely, when triggering conditions or context change, agents reconsider intentions and may drop, suspend, or replace them. Mapping this to modern GenAI-centred agents is straightforward and useful:

\textbf{Beliefs → world state and memory}. This includes short-term scratchpads, episodic case logs, and long-term semantic stores (e.g., retrieval corpora). Reliability hinges on provenance, freshness, and conflict resolution within this memory.

\textbf{Desires → goals and constraints}. Goals originate from users, alerts, or policies; constraints embed safety limits, budgets, and compliance rules. Reliability requires goal normalisation (unambiguous, measurable objectives) and policy checking.

\textbf{Intentions → active plans and tool calls}. In GenAI agents, intentions manifest as structured plans, queued tool invocations, or workflows. The architectural analogue of “commitment” is a guarded execution pipeline: once an intention is adopted, the system proceeds through validated steps unless a reconsideration trigger fires (e.g., new evidence, violated pre-conditions, exceeded risk).

BDI also clarifies where to place assurance, since \textbf{adoption filters} decide when a candidate goal becomes an intention (e.g., after feasibility checks or human approval), \textbf{execution monitors} track whether context still supports the intention (sensor deviations, failed pre-conditions, policy conflicts), and \textbf{reconsideration policies} specify how often and under what signals the agent re-plans, trading stability against responsiveness.

Finally, BDI helps articulate the boundary with generative components. Generative models may propose beliefs (summaries, hypotheses) and draft plans, but the \textbf{agent architecture} must \textit{adopt, commit, monitor, and revise} them according to explicit rules. This separation turns free-form generation into governed behaviour: intentions are treated as contracts with pre-conditions, invariants, and post-conditions; memory updates carry typed structures and provenance; and re-planning is budgeted and auditable. In short, classical foundations supply the control skeleton that keeps modern agentic systems reliable when facing latency, brittleness, and partial observability.

\section{A taxonomy of modern agentic architectures}

\begin{figure}[t]
    \centering
    \begin{tikzpicture}[
  font=\scriptsize,
  node distance=2.2mm and 1mm,
  FamHeader/.style={draw, rounded corners, thick, fill=black!10,
                    align=center, inner sep=0mm, text width=15mm,
                    minimum height=8mm},
  FamRow/.style={draw, rounded corners, thick, fill=black!5,
                 align=center, inner sep=0mm, text width=15mm,
                 minimum height=15mm},
  CellHeader/.style={draw, rounded corners, fill=black!5,
                     align=left, inner sep=2mm, text width=15mm,
                     minimum height=8mm},
  CellRow/.style={draw, rounded corners,
                  align=left, inner sep=1mm, text width=17mm,
                  minimum height=8mm},
  gridline/.style={draw=black!15, line width=0.3pt}
]

% --- Header row (h1..h6) ---
\node[FamHeader]                        (h1) {Family};
\node[CellHeader, right=1mm of h1]      (h2) {Reasoning \&\\ control};
\node[CellHeader, right=1mm of h2]      (h3) {Memory};
\node[CellHeader, right=1mm of h3]      (h4) {Action surface};
\node[CellHeader, right=1mm of h4]      (h5) {Assurance\\ envelope};
\node[CellHeader, right=1mm of h5]      (h6) {Risks $\rightarrow$\\ mitigations};

% === Row 1: Tool-using (anchor on rightmost) ===
\node[CellRow, below=3mm of h6]         (r1c6) {Hallucinated tools, overreach $\rightarrow$ allow-list, least privilege, tool mocks};
\node[CellRow, left=1mm of r1c6]        (r1c5) {Allow-lists; \\validate schema; idempotent retries; rate limits};
\node[CellRow, left=1mm of r1c5]        (r1c4) {APIs \& tools via router; \\typed schemas};
\node[CellRow, left=1mm of r1c4]        (r1c3) {Scratchpad; \\keep minimal state only};
\node[CellRow, left=1mm of r1c3]        (r1c2) {Prompted CoT; reactive steps; shallow plans};
\node[FamRow,  left=1mm of r1c2]        (r1c1) {Tool-\\using};

% === Row 2: Memory-augmented ===
\node[CellRow, below=3mm of r1c6]       (r2c6) {Poisoned or stale context $\rightarrow$ recency filters, provenance checks, dedupe};
\node[CellRow, left=1mm of r2c6]        (r2c5) {Provenance; confidence thresholds; \\freshness \& \\recency checks};
\node[CellRow, left=1mm of r2c5]        (r2c4) {Read \& write notes; cite \& summarise sources};
\node[CellRow, left=1mm of r2c4]        (r2c3) {Vector store; episodic logs; entity KB};
\node[CellRow, left=1mm of r2c3]        (r2c2) {Retrieval-augmented\\ reasoning; \\query planning};
\node[FamRow,  left=1mm of r2c2]        (r2c1) {Memory-augmented};

% === Row 3: Planning / self-improvement ===
\node[CellRow, below=3mm of r2c6]       (r3c6) {Spec gaming \& loops $\rightarrow$ reward shaping, step caps, reviewers};
\node[CellRow, left=1mm of r3c6]        (r3c5) {Sandboxing; test gates; budget \& termination codes};
\node[CellRow, left=1mm of r3c5]        (r3c4) {Code execution; unit tests;\\ self-play};
\node[CellRow, left=1mm of r3c4]        (r3c3) {Execution traces; model edits; curricula};
\node[CellRow, left=1mm of r3c3]        (r3c2) {Deliberate multi-step \\ planners;\\ self-critique};
\node[FamRow,  left=1mm of r3c2]        (r3c1) {Planning \& self-improve};

% === Row 4: Multi-agent ===
\node[CellRow, below=3mm of r3c6]       (r4c6) {Coordination failure, chatter $\rightarrow$ protocols, quorum, leadership hand-off};
\node[CellRow, left=1mm of r4c6]        (r4c5) {Turn-taking; moderation; \\audit trails};
\node[CellRow, left=1mm of r4c5]        (r4c4) {Messaging; tool calls; negotiation protocols};
\node[CellRow, left=1mm of r4c4]        (r4c3) {Shared board; per-agent episodic \\memory};
\node[CellRow, left=1mm of r4c3]        (r4c2) {Role-specialised agents; auctions \& blackboard; coordinator};
\node[FamRow,  left=1mm of r4c2]        (r4c1) {Multi\\-agent};

% === Row 5: Embodied / web ===
\node[CellRow, below=3mm of r4c6]       (r5c6) {Irreversible actions with side-effects $\rightarrow$ kill-switch, simulate-before-actuate, permits};
\node[CellRow, left=1mm of r5c6]        (r5c5) {Simulation; safety envelope; human-in-the-loop};
\node[CellRow, left=1mm of r5c5]        (r5c4) {Web automation or robots; browsers or physical actuators};
\node[CellRow, left=1mm of r5c4]        (r5c3) {World model; maps; caches};
\node[CellRow, left=1mm of r5c3]        (r5c2) {Perception $\rightarrow$ action loop;\\ reactive planners};
\node[FamRow,  left=1mm of r5c2]        (r5c1) {Embodied or web};

% --- Grid lines (optional) ---
% \begin{scope}[on background layer]
%   % vertical lines between columns 2–5 (use first/last row cells)
%   \foreach \Top/\Bot in {r1c2/r5c2, r1c3/r5c3, r1c4/r5c4, r1c5/r5c5}
%     \draw[gridline] (\Top.north west) -- (\Bot.south west);
%   % horizontal lines between family rows
%   \foreach \L/\R in {r1c1/r1c6, r2c1/r2c6, r3c1/r3c6, r4c1/r4c6, r5c1/r5c6}
%     \draw[gridline] (\L.south west) -- (\R.south east);
% \end{scope}

\end{tikzpicture}
    \caption{Comparison of agentic architecture families across core design facets, showing how each approach shapes capabilities, risks, and the reliability tools required to control behaviour.}
    \label{fig2}
\end{figure}
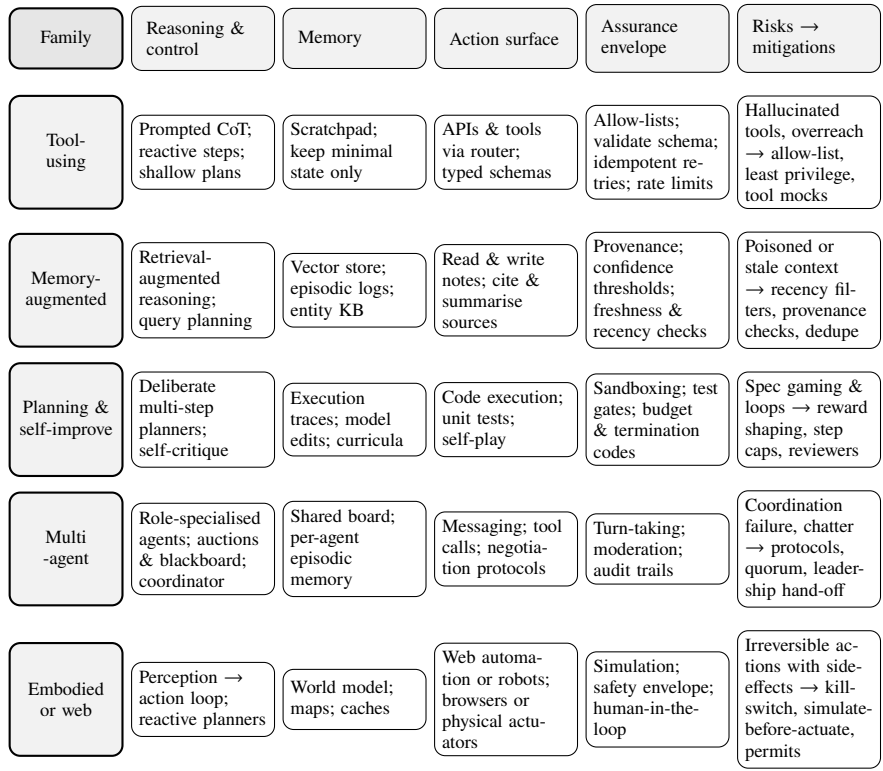

Modern agentic systems are not single, monolithic designs, but rather compositions of different building blocks—planning loops, tool routers, memory layers, verifiers, and supervisors—typically stitched together in various ways around a core of generative reasoning. A useful taxonomy, therefore, needs to be \textbf{pattern-centric}, not model-centric: it should explain common patterns of how these building blocks combine, what interfaces bind them, and how those choices affect reliability. The goal of this section is to situate the reader in that design space before we examine selected patterns in more depth.

We organise the taxonomy along a small set of \textbf{orthogonal facets} that cut across implementations: (i) \textit{reasoning mode} (reactive prompts, plan-then-act, interleaved reasoning-and-acting, search over thoughts/programs); (ii) \textit{memory strategy} (stateless, scratchpad, episodic, semantic/RAG, long-term embodied memory) and the associated provenance and hygiene controls; (iii) \textit{control structure} (single agent, hierarchical supervisor–worker, or peer multi-agent with protocols); (iv) \textit{action surface} (read-only tools, simulators/digital twins, or actuators with real-world side-effects) and its permissioning; and (v) \textit{assurance envelope} (validators, critics/verifiers, runtime supervisors, and fallback controllers). These facets should, ideally, be \textbf{composable}: a system may, for example, combine interleaved reasoning-and-acting with RAG memory, a supervisor–worker team topology, and strict schema-validated tool calls.

Within this framing, we group contemporary designs into five families that recur in practice and are easiest to teach: \textbf{tool-using agents, memory-augmented agents, planning and self-improvement agents, multi-agent systems, and embodied or web agents}. For each family, we will emphasise the reliability perspective and aspects it naturally affords—controllability, observability, recoverability, and governability. We also discuss the typical failure modes it is particularly susceptible to. Readers can map these families onto the running example (the tool-using diagnosis agent) to see how swapping a pattern—e.g., adding a critic, transitioning from single- to multi-agent, or upgrading memory from scratchpad to RAG—changes the assurance story without requiring a complete system rewrite.

\subsection{Tool-using agents}

Tool-using agents place a generative model in charge of \textbf{deciding which external capabilities to invoke, with what arguments, and in what order}, to accomplish a goal. The spectrum ranges from simple \textit{function calling} (where the model emits a structured payload that a tool executes) to \textit{tool learning}, where the model learns—often from weak supervision—\textbf{when} to call an API and \textbf{how} to incorporate the results into its subsequent reasoning. The latter was demonstrated by Toolformer~\cite{Toolformer}, which trains a language model to self-label API calls from a few demonstrations and thereby improves performance without sacrificing general language ability. This shift—from “LLM answers directly” to “LLM orchestrates instruments”—is the core move that turns a predictor into an \textbf{agent}.

A robust implementation separates three concerns. \textbf{Planners} generate \textit{intentional structure} (sub-goals, hypotheses, candidate actions). \textbf{Tool routers} map these abstract actions to concrete tools, select among alternatives, and fill arguments using typed schemas and retrieval. \textbf{Execution sandboxes} then perform pre-condition checks, rate-limit and permission each call, simulate where possible, and commit side-effects transactionally (or roll them back). This decomposition, as we mentioned before, localises faults: planners may hallucinate, but routers refuse ill-typed or out-of-policy calls; sandboxes contain side-effects and provide deterministic replay. In practice, you gain levers for reliability at each boundary: schema validation and capability scoping at the planner–router interface; idempotency keys, timeouts, and compensating actions at the router–sandbox interface; and comprehensive, structured traces across all three.

For obvious reasons, this chapter cannot be comprehensive in describing all the available architectures. Therefore, we have somewhat subjectively selected three influential design patterns to illustrate distinct trade-offs:

\textbf{MRKL (Modular Reasoning, Knowledge and Language)}~\cite{MRKL} argues for \textit{neuro-symbolic modularity}: keep the language model for flexible decomposition and natural-language control, but route specific sub-tasks (search, calculator, database queries, discrete reasoning) to specialised modules. Architecturally, MRKL is a \textit{tool-router-first} worldview: the LLM is not a monolith but a foreman that dispatches to curated tools and knowledge sources. Reliability benefits come from explicit module contracts (clear I/O shapes, error codes) and from narrowing each tool’s authority; the cost is integration complexity and the need to manage versioned tool APIs and their provenance.

\textbf{ReAct (interleaved reasoning + acting)}~\cite{ReAct} prompts the model to alternate between \textit{thought} (chain-of-thought style reasoning traces) and \textit{action} (tool calls), allowing observations to feed back immediately into subsequent reasoning. This yields human-readable trajectories and can reduce hallucinations because evidence is actively fetched rather than imagined. The reliability angle is twofold: (i) interpretable traces make it easier to \textit{audit and intervene}; (ii) the tight loop can still run away without limits, so one must bound step counts, enforce per-tool budgets, and validate every action before execution. In safety-critical settings, a \textit{ReAct with a governor} is recommended, as it requires that each proposed action is schema-checked, policy-checked, and—where feasible—simulated before it is allowed to affect the real world.

\textbf{ReWOO (Reasoning Without Observation) Planner–Executor}~\cite{ReWOO} \textit{decouples} the generation of a complete reasoning plan from the acquisition of observations. The model first drafts a symbolic plan referencing place-holders for tool outputs; only then does the system execute the required tool calls to fill those slots, followed by a light-weight synthesis step. This separation reduces token churn and latency from repeated “think–act–think” cycles, and the plan itself becomes an auditable artefact with verifiable pre- and post-conditions. Reported evaluations show substantial token efficiency gains and improved robustness under tool failures, precisely the regimes that matter in real-world deployment. The reliability gain is architectural: plans are \textit{objects}—they can be statically checked, costed, approved by a supervisor, and even cached or replayed—before any side-effectful call occurs.

Across these patterns, the \textbf{assurance envelope} hinges on a relatively small set of key engineering choices:
\begin{itemize}
\item \textit{Typed schemas and validators.} Treat every tool invocation as a contract (e.g., JSON Schema). Reject on mismatch; surface structured errors back to the planner. This alone eliminates a large class of silent failures and injection-style exploits where model outputs smuggle unintended arguments.
\item \textit{Idempotent tools and request deduplication.} Side-effectful tools should accept idempotency tokens and be safe under retries; read paths should be pure. This makes recovery from partial failures predictable and supports exactly-once semantics at the orchestration layer.
\item \textit{Capability-based permissioning.} Bind tools to least privilege (read-only by default; actuation only for narrow, reversible operations), with per-call policy checks that incorporate user, context, and risk. In an incident, the constraining of authority limits the blast radius.
\item \textit{Transactional semantics and compensations.} Where ACID is unavailable (e.g., multi-service actions), use “saga” patterns: write-ahead intent logs, outbox delivery, and compensating actions. Pair these with simulation-before-actuation for high-risk calls.
\item \textit{Budgets and termination criteria.} Enforce step limits, cumulative cost caps, and wall-clock timeouts at the orchestration layer; emit “why-stopped” codes for auditability and post-mortems.
\item \textit{Deterministic observability.} Log plans, tool arguments, results, and decision rationales in structured form; include hashes of inputs, tool versions, and policy snapshots so you can replay and explain outcomes.
\end{itemize}

Finally, there is a need to consider \textbf{failure modes} characteristic of tool-using agents and their mitigations. One common challenge is \textit{hallucinated tools or arguments}, which can be addressed by whitelisting tool names, validating data against schemas, and requiring router approval. Of course, there is a tradeoff between reliability and creativity in constraining the use of tools available to the agent. \textit{Infinite or unproductive loops} are another danger, where imposing clear steps or time budgets with safe halts can be necessary. Determining the exact thresholds, however, is not necessarily straightforward. Consideration for \textit{tool flakiness} requires a ``design for retry'' mentality with jitter, circuit breakers, and degrade to read-only advice. A clear danger is \textit{prompt or retrieval injection}, which requires sanitising tool outputs, stripping control tokens, and treating any untrusted text as data, not instructions—all easier said than done. The guiding principle is straightforward, even if far from trivial to implement robustly: \textbf{models propose, architectures dispose}—with contracts, governors, and sandboxes that convert open-ended reasoning into reliable action.

\subsection{Memory-augmented agents}

Agentic systems become markedly more capable when they can \textbf{remember}: carry forward intermediate reasoning (working memory), accumulate case-specific experience (episodic memory), and ground answers in stable, queryable knowledge (semantic memory). These three layers serve distinct functions, yet they also work in synergy with one another. 

\textbf{Working memory} is a short-lived context the model manipulates within an episode—scratchpads, intermediate steps, and \textit{chain-of-thought} (CoT)~\cite{CoT} traces—which improve decomposition, error checking, and tool selection.

\textbf{Long-term memory} can be accessed on demand, and splits into (i) \textit{episodic} stores (structured logs of prior tasks, decisions, tool invocations, and outcomes) and (ii) \textit{semantic stores} (documents, facts, tables). As an example, \textit{Retrieval-Augmented Generation} (RAG)~\cite{RAG} provides a non-parametric pathway into semantic memory: instead of relying solely on weights, the agent retrieves passages from an index and conditions generation on them, improving factuality and enabling updates without retraining. In the original formulation, RAG combines a pretrained ``seq2seq'' model with a dense retriever over a corpus (e.g., Wikipedia), yielding more specific and verifiable answers on knowledge-intensive tasks.

As contexts and histories grow, \textbf{memory management} itself becomes an \textit{architectural concern}. One recent influential pattern is \textit{OS-style virtual context}: the agent maintains a small, fast ``working set'' inside the model’s context window and pages additional information in and out from external stores, guided by control signals. \textbf{MemGPT}~\cite{MemGPT} exemplifies this: it orchestrates multi-tier memory (fast context vs slower external stores), uses “interrupts” to govern control flow, and automatically retrieves or evicts content so the model can operate over effectively unbounded histories despite a finite context window. This reframes memory from a passive store to an \textit{active subsystem} with policies for admission, eviction, and prefetch.

Designing memory for \textbf{reliability}, not merely performance, requires explicit contracts:

\textbf{Provenance and attribution}. Every retrieved or persisted item should carry source identifiers (URI or document ID), a content hash, a timestamp, and the retrieval policy/version that produced it. Plans, diagnoses, or actuation proposals must cite the memory elements on which they depend; logs should retain these bindings so that episodes are replayable and audit trails are intact. In RAG pipelines, this means storing the retriever configuration, top-k, similarity scores, and any re-ranking steps alongside the selected passages. Provenance is the first defence against spurious correlations and a prerequisite for compliance and post-incident analysis.

\textbf{Freshness and validity windows}. Not all knowledge ages equally. Introduce \textit{time-to-live} and \textit{refresh-on-access} policies by source, with staleness thresholds that trigger re-indexing or explicit human review. For episodic memories (e.g., prior interventions on a machine), encode validity conditions—such as firmware version, environment, or configuration—so retrieval is \textit{conditional} on compatibility, thereby avoiding dangerous ``near matches''. In virtual-context systems, freshness policies should influence paging priority to ensure up-to-date items dominate the reasoning.

\textbf{Memory hygiene and poisoning defences}. Treat all untrusted text (web pages, user input, third-party tool output) as \textit{data}, never as executable control. Apply sanitisation (e.g., strip model-control tokens), restrict which fields can flow into prompts, and use allow/deny lists for pattern-based blocking. Guard the write path: only curated processes may persist to long-term stores; user-generated content goes to quarantine until vetted. Maintain \textit{trust tiers} (gold, silver, untrusted) that affect retrieval ranking and whether citations are mandatory. Align threats with recognised taxonomies (prompt injection, insecure output handling, data poisoning) and test with adversarial corpora~\cite{OWASP}.

\textbf{Compaction, summarisation, and retention}. Long-running agents accumulate vast episodic traces. Without compaction, retrieval degrades and costs rise. Use layered retention: keep lossless structured logs indefinitely in cold storage; maintain \textit{summarised} episodic memories (task → actions → outcomes → lessons) for mid-term recall; and cache \textit{salient} spans in hot stores for rapid access. Summaries should be \textit{checked against sources} (verifier pass) and carry back-pointers to the raw logs. Eviction should be policy-driven (LRU, recency-frequency, or task-aware scoring), and compaction jobs must be idempotent and versioned so that summaries can be regenerated deterministically. MemGPT-style pagers can use these scores to drive what remains in the active working set.

\textbf{Separation of concerns in the memory stack}. Keep \textit{working memory} (scratchpads/CoT) isolated from \textit{long-term stores} to prevent leakage of speculative thoughts into durable knowledge. Working memory improves reasoning but is intentionally disposable; CoT artefacts should not be permitted to write directly to semantic stores without a gate (e.g., a tool-mediated ``knowledge write'' that demands evidence and passes a verifier). Conversely, RAG retrievals should enter prompts via \textit{typed slots} (title, snippet, citation) with explicit delimiters to limit injection. This insulation clarifies which errors are reasoning slips versus memory contamination.

From an \textbf{engineering} standpoint, a reliable memory subsystem exposes \textit{deterministic APIs}: 
\begin{itemize}
    \item \verb|retrieve(query, policy) → {items, scores, policy_id}|
    \item \verb|write(record, policy) → {id, version}|
    \item \verb|page_in(keys)|
    \item \verb|evict(keys, reason)|
\end{itemize}

It logs decisions (why this item ranked; why that summary replaced these events), and it ships with \textit{self-tests}: poisoning probes, staleness alarms, and retrieval drift monitors that detect when changes in embeddings, corpus, or policies shift behaviour. For safety-critical agents, it is recommended to add a \textit{two-phase write} to semantic stores (draft → verified → published) and require corroboration from multiple sources before knowledge is considered ``actionable''.

Finally, it is crucial to remember that memory choices \textbf{shape system behaviour over time}. Well-designed episodic stores enable agents to self-improve (they learn from what worked last time), whereas poorly governed stores tend to calcify early mistakes. RAG broadens competence but expands the attack surface; virtual context rescues long-horizon tasks but introduces paging pathologies if policies are naïve. The remedy is architectural discipline: treat memory as a first-class subsystem with provenance, freshness, hygiene, and retention policies—not as a bag of vectors appended to a prompt.

\subsection{Planning- and self-improvement agents}

Planning-centred agents strengthen a model’s raw reasoning with \textbf{explicit search, external executors, and self-evaluation loops}. The core architectural move is to separate \textit{proposal} (generate candidate thoughts, plans, or programs) from \textit{selection} (score and prune) and \textit{execution} (run code, query tools, or act). We will use four families to exemplify these ideas.

\textbf{Tree of Thoughts (ToT)}~\cite{ToT} organises inference as a \textit{search over intermediate thoughts}, rather than a single left-to-right pass (as in CoT). The agent expands a tree whose nodes are candidate partial solutions; a scorer (often the model itself or a light critic) evaluates nodes, while the controller chooses which branches to expand or backtrack. In effect, ToT trades tokens for \textit{look-ahead} and \textit{global choice}, enabling recovery from early local mistakes. Reported case studies (e.g., Game of 24, creative tasks, mini crosswords) show large gains over plain chain-of-thought by exploring multiple reasoning paths and permitting backtracking. Architecturally, ToT exposes explicit levers—branching factor, depth, and scoring policy—that you can govern and log, making the agent’s computation more \textit{auditable} and \textit{replayable}.

\textbf{Graph of Thoughts (GoT)}~\cite{GoT} generalises ToT from trees to \textit{arbitrary dependency graphs}. Thoughts become vertices, and edges record informational dependencies, allowing for the recombination, summarisation, or refinement of subgraphs and \textit{feedback loops} across them. This added flexibility is particularly important for tasks where sub-problems interlock (e.g., sorting with constraints, multi-step data transformations). Graphs permit the reuse of partial results and parallel exploration while still retaining a programmable controller that shapes the search. Empirical results demonstrate both \textit{quality improvements} and \textit{cost reductions} compared with ToT by sharing and distilling subgraphs rather than re-deriving them along separate branches. For reliability, GoT’s graph abstraction offers natural checkpoints—subgraphs can be validated, cached, and reused—and clearer provenance: each final answer can cite the exact subgraph it depends on.

\textbf{Program-Aided Language models (PAL)}~\cite{PAL} shift the heavy lifting from natural-language reasoning to \textit{executable programs} generated by the model and run by a trusted interpreter. The model reads a problem, emits code (e.g., Python) that embodies the reasoning, and an external runtime executes it to produce the answer. PAL routinely outperforms larger models relying on free-form chain-of-thought for algorithmic and mathematical tasks because interpreters provide precise, deterministic computation and a crisp failure mode (exceptions) rather than silent arithmetic drift. In other words, PAL converts ambiguous ``reasoning text'' into \textit{a specification that the machine can run}, which is far easier to validate, sandbox, and test.

\textbf{Reflexion}~\cite{Reflexion} \textbf{adds self-evaluation and test-time repair} without gradient updates. After an attempt, the agent produces a \textit{verbal reflection}—a concise diagnostic of what failed—and stores it in episodic memory. On the next trial, this reflection steers prompts, search, or program synthesis, yielding rapid improvements across sequential decision-making, coding, and reasoning tasks. Importantly, Reflexion transforms feedback (scalar rewards, failures, or critiques) into \textit{structured guidance} that persists across episodes, facilitating improvements that are architectural rather than purely parametric.

From a reliability standpoint, planning and self-improvement agents are powerful but also potentially quite unruly. The architecture must therefore centre on \textit{search control}, \textit{verification}, and \textit{cost governance}.

\textbf{Search control}. Expose and enforce branching, depth, and expansion policies (beam width, temperature schedules, stopping rules). For ToT/GoT, prefer \textit{budget-aware controllers}: allocate a fixed token/time budget and use adaptive pruning (e.g., softmax over scores with a floor on exploration) so search cannot run away. Record \textit{why-stopped} codes (budget exhausted, convergence, contradiction) to aid post-mortems. For tasks with heterogeneous hardness, a \textit{bandit-style scheduler} can allocate compute to promising branches while guaranteeing exploration quotas.

\textbf{Verifiers and critics}. Decouple proposing from judging. Use \textit{domain verifiers} (unit tests for PAL programs; invariants for plans; checkers for constraint satisfaction) before accepting a branch. Keep critics \textit{tool-assisted}: combine LLM critique with symbolic checks (type, range, satisfiability). In GoT, validate subgraphs as independent artefacts and reuse only those that pass. Treat the verifier as a \textit{trust boundary} that must be deterministic, sandboxed, and versioned.

\textbf{Test-time repair}. When verifiers fail, embed repair strategies: patch programs (PAL) guided by error traces; revise subgraphs (GoT) using counter-examples; or inject Reflexion-style \textit{failure summaries} into the next proposal phase. Log each repair attempt and link it to the failing artefact so the system accumulates \textit{case-based fixes} that are auditable.

\textbf{Cost and latency governance}. Planning expands computation non-linearly. Enforce Service Level Objectives with per-request caps on tokens, wall-clock, and tool calls; pre-empt long-running searches; and favour \textit{anytime behaviours} (best-so-far answer with a quality estimate). Cache validated partials (subtrees/subgraphs, passing PAL tests) under input fingerprints to avoid recomputation. Track \textit{cost-to-success} and \textit{retry counts} as first-class metrics.

\textbf{Sandboxed execution}. For PAL and code-generating variants, run interpreters in \textit{constrained sandboxes} (no network, bounded CPU/memory/time, whitelisted packages). Prefer pure functions; where state is unavoidable, add idempotency and snapshot/rollback.

\textbf{Provenance and replay}. Treat search as a dataset. Persist node/edge contents, scores, verifier outputs, and controller decisions with hashes of prompts, model/version, and policies. This enables \textit{deterministic replay}, fault isolation (proposal vs verifier vs controller), and targeted regression tests when upgrading components.

Typical failure modes and mitigations follow naturally. \textit{State explosion} can be addressed by budgeted expansion with admissible heuristics and early pruning. \textit{Speculative arithmetic or logic errors} call for PAL-style external execution with unit tests. \textit{Over-confident selection} is handled by a separate verifier, and disagreement checks are enforced (accept the solution only if the verifier concurs). \textit{Cost cliffs} are mitigated by anytime controllers and cached subgraphs.

In summary, planning- and self-improvement agents deliver \textbf{substantial reliability dividends} when their power is channelled through explicit controllers, trustworthy verifiers, and disciplined governance of cost and side-effects. ToT and GoT make reasoning \textit{searchable and auditable}; PAL makes it \textit{executable and testable}; Reflexion makes it \textit{learnable at inference time}. Together, they convert raw generative competence into \textbf{governed problem-solving} that improves with experience rather than drifting unpredictably.

\subsection{Multi-agent systems}

Multi-agent systems exchange a single ``do-everything'' agent for a \textbf{team of specialised agents} that co-operate (or compete) under explicit protocols. In practice, three topologies have emerged. \textbf{Supervisor–worker} designs place the coordinator in charge of task decomposition, assignment, and arbitration; they are easy to govern because authority and escalation paths are clear. \textbf{Peer collaboration} removes the central coordinator and relies on protocol rules (e.g., turn-taking, proposal–critique–revise cycles) to drive convergence. \textbf{Role-play protocols} script complementary roles (e.g., domain expert vs. software engineer), using \textit{inception prompts} to maintain role consistency and reduce drift.

Frameworks such as AutoGen~\cite{AutoGen} treat conversation itself as the computation, allowing developers to program interaction graphs (who talks to whom, with which tools, and when to stop), while CAMEL~\cite{CAMEL} demonstrates that stable roles and carefully seeded goals can produce reliable cooperative behaviour without a human in the loop.

A protocol is more than turn-taking; it encodes \textit{who may propose, who may critique, what constitutes evidence}, and \textit{how decisions are adopted}. In supervisor–worker teams, the supervisor enforces work allocation, deadlines, and acceptance tests, and may reassign or down-scope tasks in the event of failure. In peer teams, robustness depends on \textit{interaction motifs} such as (i) proposal → cross-examination → revision; (ii) debate → referee verdict; or (iii) consensus vote after bounded discussion. Role-play (à la CAMEL) reduces unproductive behaviours (role flipping, repetition, infinite loops) by fixing complementary perspectives and anchoring each agent’s incentives to the shared goal. AutoGen exposes these choices as first-class: developers specify agent roles, tool permissions, conversational edges, and stop conditions, making the interaction configurable, replayable, and inspectable.

Reliability is primarily controlled by five aspects:

\textbf{Protocol invariants}. Treat messages as typed artefacts: require schemas, citations for claims, and explicit \textit{proposal/critique/decision} labels; forbid agents from escalating authority or mutating roles at run-time. For tool-enabled agents, attach capability scopes to roles (e.g., only the ``Executor'' may call actuation tools, and only after the ``Verifier'' approves). CAMEL explicitly documents recurrent failure patterns (role flipping, ``flake'' replies, infinite loops), which these invariants aim to prevent.

\textbf{Termination conditions}. Multi-agent dialogues easily diverge. Enforce \textit{hard} limits (rounds, wall-clock, cumulative token/cost budgets) and \textit{soft} stop rules (no-new-information, fixed-point detection, repeated proposals). Every dialogue should finish with a \textit{why-stopped} code (e.g., ``consensus reached'', ``budget exceeded'', ``non-convergence'').

\textbf{Arbitration}. When agents disagree, appoint an \textit{arbiter} (a supervisor, a specialised ``Referee'' agent, or a symbolic verifier). Define tie-breakers (confidence thresholds, external tests, human escalation) and keep the arbiter’s decision rule deterministic and versioned.

\textbf{Consensus/critique loops}. Debate improves accuracy but can entrench sycophancy (agents aligning on a wrong answer). Use heterogeneous critics (different prompts/models), require \textit{evidence-backed} critiques, and gate adoption through verifiers or a referee. Recent work on consensus-seeking multi-agent debates shows accuracy gains when agreement bias is explicitly mitigated and when the number of rounds is optimised for cost~\cite{CONSENSAGENT}.

\textbf{Failure isolation}. Run agents in \textit{separate sandboxes} with least-privilege tools; propagate only \textit{summaries} or \textit{typed facts} between agents, not raw prompts; and checkpoint intermediate artefacts (plans, code, proofs) so that a faulty agent can be restarted or replaced without losing global progress.

Engineering the conversation-as-computation layer can be done by making the conversation engine an explicit subsystem with its own APIs and telemetry. Each exchange should capture the \textit{role, speech-act, payload, evidence, tool-calls, and decision-flag}, plus hashes of prompts, model versions, and policies. Auto-generated \textit{conversation DAGs} (with nodes representing messages or artefacts and edges representing dependencies) enable \textit{subset replay}, targeted audits, and caching of validated sub-results for later reuse. Supervisors should enforce \textit{per-role budgets} and \textit{per-edge rate limits} to prevent chat storms. Where agents emit code or executable plans, they require \textit{pre-commit tests} (such as unit tests and static checks) and \textit{post-commit monitors} (such as runtime invariants) before any side-effectful actuation.

Typical failure modes include runaway dialogues or deadlocks, echo-chamber agreement bias, role drift with unauthorised authority escalation, and uncontrolled error propagation across participants. Mitigate these with hard round/time caps and deadlock detection plus supervisor pre-emption; diversify critics and use blinded proposals with a referee that scores evidence and mandates a disagreement phase; enforce immutable role descriptors with capability tokens and schema checks; and quarantine low-trust outputs, require dual control for high-impact actions, and equip each agent with circuit breakers.

The guiding principle is to \textbf{treat the team protocol as the primary object of assurance}. AutoGen’s programmable interaction graphs and CAMEL’s role-stable dialogues provide the raw mechanisms; reliability comes from adding invariants, budgets, arbitration, and isolation so that a group of powerful but fallible agents behaves like a disciplined organisation rather than a noisy crowd.

\subsection{Embodied or web agents}

Embodied and web agents act \textbf{in the world} rather than merely \textbf{about it}. The former couple decisions to sensors and actuators in physical environments (robots, autonomous platforms, industrial control); the latter operate browsers, APIs, and GUIs across \textit{untrusted}, \textit{changing} websites and enterprise systems. Both classes face amplified risks: (i) \textit{unbounded inputs} (sensor noise, adversarial pages, altered layouts); (ii) \textit{long-horizon side-effects} (actions whose consequences accumulate or branch widely); and (iii) \textit{opaque dependencies} (firmware or toolchains for embodied agents; third-party scripts, cookies, and authentication flows for web agents). Accordingly, they demand stricter \textbf{runtime assurance}, \textbf{sandboxing}, and \textbf{governance} than purely analytical agents. For both embodied and web settings, actuation must be treated as a \textit{privileged boundary}. It is crucial to \textbf{simulate-before-actuate} and route proposed actions through a digital twin (physical) or a dry-run/snapshot DOM (web) to check pre-conditions, invariants, and expected deltas. Require a verifier’s \textit{green light} before issuing any irreversible command. \textbf{Least-privilege capability tokens} should be used to bind actions to narrow, time-limited capabilities (e.g., ``move $\leq$ 0.2 m at $\leq$ 0.1 m/s'' or ``HTTP GET on whitelisted domains only''). Deny escalation at run-time. Finally, \textbf{supervisors and safe fallbacks} must enforce budget and risk limits; provide \textit{graceful degradation} (monitor-only, read-only, or ``shadow mode'' that mirrors actions without effect). Make interruption and rollback first-class: every critical action should be reversible or followed by a compensating step.

Physical systems add dynamics and contact physics to the reasoning problem. Architectures should \textbf{separate fast safety from slow reasoning}. Keep protective control (interlocks, torque/velocity limiters, geofences) in a fast loop near the plant; place planning, task logic, and learning in supervisory layers. The safety layer enforces hard constraints regardless of the planner’s output. \textbf{Certified monitors} should continuously ensure key invariants, implementing control barrier functions, speed/force envelopes, and zone exclusion as deterministic guards; violations must be logged and trip to safe states. 

Web agents primarily need to consider \textbf{security-first interaction} with \textbf{volatile UIs}. Web automation expands the attack surface as DOMs change, scripts execute, and adversarial prompts can be embedded in pages or PDFs. Hardened browsers running in \textbf{sandboxes} and isolated containers without general network access; dangerous APIs (downloads, file system, microphone/camera) should be disabled, together with third-party extensions, and uncontrolled script execution. Deterministic \textbf{interaction contracts} are strongly preferred (e.g., stable API calls over brittle UI scraping); if browsing is necessary, pin to \textit{selectors with checksums} (element text/structure hashes) and version layouts. Treat all page text as \textit{data}, never as executable control; sanitise and label untrusted content before it enters prompts. Allow/deny lists and transaction guards should include whitelisting domains and HTTP verbs; require \textit{dual control} (human or verifier) for POST/PUT/DELETE, payments, or credential flows. Persist a \textit{pre-commit diff} (what will change) and obtain approval before execution. For \textbf{authorisation and secrets}, always use short-lived tokens bound to scope and device; never expose secrets to the model; inject credentials at the sandbox boundary, not in prompts.

In short, embodied and web agents convert open-ended reasoning into \textbf{world-altering behaviour}. Their reliability is directly tied to the architecture: simulate before actuate, gate authority with narrow capabilities, enforce deterministic guards at the boundary, and instrument everything for replay.

\section{Architectural building blocks and interfaces}

At the level of components, dependable agentic systems tend to converge on the same spine: a \textbf{goal manager} to normalise objectives and constraints; a \textbf{planner} to propose decompositions and candidate actions; a \textbf{tool-router} to map abstractions to concrete capabilities; an \textbf{executor} (with sandboxing) to perform calls; \textbf{memory} layers (episodic logs and semantic/RAG stores) to ground decisions; \textbf{verifiers/critics} to check plans, results, and policies; a \textbf{safety monitor} to enforce budgets, invariants, and fallbacks; and \textbf{telemetry} that makes all of this observable and replayable. This separation of concerns provides natural choke points for validation, authority, and audit.

Reliability is then carried by \textbf{structured interfaces}. Prefer \textit{function schemas} and \textit{structured outputs} so that model proposals must conform to a declared JSON schema; this turns free-form text into typed objects and sharply reduces silent failures in downstream tooling. Pair schema enforcement with \textit{validators/guardrails} (type, range, policy, content filters) and bind all tools to \textit{capability-scoped permissions} (least privilege, time-limited tokens, idempotency). Taken together, these choices make actions either ``deterministically safe'' or ``deterministically refused''~\cite{JSON}. 

For \textbf{orchestration}, two families cover most needs. LangGraph~\cite{LangGraph} provides a low-level, stateful graph runtime: you define a shared state, nodes that transform it, and edges that encode control flow—ideal when you need explicit, replayable workflows and tight governance over long-running agents. AutoGen~\cite{AutoGen2} treats conversation as computation for single- and multi-agent settings: you program roles, message routes, tool access, and stop conditions—useful when collaboration, debate, or supervisor–worker patterns are central. In both cases, constrain them with the same contracts: typed messages, per-edge budgets, per-role capabilities, and mandatory verifiers before any side-effectful act.

If the priority is \textit{deterministic control and auditability} of a single agent (or a small number of tightly coupled ones), start with a state-graph (e.g., LangGraph) and insist on schema-checked nodes and transactional executors. If your priority is \textit{division of labour and critique} across multiple roles, use a conversation framework (e.g., AutoGen) but add protocol invariants, termination rules, and an arbiter/verifier at the boundary to the real world. In either case, keep schema enforcement and validators close to the tool boundary, and record hashes of state, prompts, policies, and tool versions for replay.

The \textbf{key takeaway message} is that architecture—not just model choice—determines whether a powerful reasoning core becomes a dependable system. The components above provide isolation; schemas and validators provide contracts; orchestration provides controlled evolution over time. Put differently: \textit{models propose, architectures dispose}. With this scaffold in place, the following chapters can focus on the specifics of data coverage, confidence estimation, cybersecurity, monitoring, and governance—plugging neatly into the interfaces you have already designed.

\section{Summary}
In this chapter, we argued that reliability in agentic and generative AI is fundamentally architectural: it is earned through principled componentisation (goal manager, planner, tool-router, executor, memory, verifiers, safety monitor, telemetry), disciplined interfaces (schema-constrained, validated, least-privilege tool calls), and explicit control loops that supervise reasoning and action. We mapped contemporary practice into a taxonomy—tool-using agents, memory-augmented agents, planning and self-improvement agents, multi-agent systems, and embodied/web agents—showing how each pattern alters the reliability envelope and introduces distinct failure modes. The diagnosis-agent example illustrated the recurring safeguards: simulate-before-actuate, typed contracts, idempotency and transactional semantics, permissioning, budget and termination rules, and end-to-end observability for audit and replay. Finally, we highlighted orchestration choices and how to constrain them so that powerful models provide solutions which are then validated before being accepted. With this scaffold in place, the subsequent chapters on data coverage, confidence estimation, cybersecurity, operational monitoring, and governance can plug into well-defined interfaces, turning capable model stacks into dependable autonomous systems.

\bibliographystyle{alpha}
\bibliography{chapter-3}

\end{document}